# LSTM Networks for Online Cross-Network Recommendations

**Dilruk Perera** and **Roger Zimmermann**
School of Computing, National University of Singapore
dilruk@comp.nus.edu.sg and rogerz@comp.nus.edu.sg

## Abstract

Cross-network recommender systems use auxiliary information from multiple source networks to create holistic user profiles and improve recommendations in a target network. However, we find two major limitations in existing cross-network solutions that reduce overall recommender performance. Existing models (1) fail to capture complex non-linear relationships in user interactions, and (2) are designed for offline settings hence, not updated online with incoming interactions to capture the dynamics in the recommender environment. We propose a novel multi-layered Long Short-Term Memory (LSTM) network based online solution to mitigate these issues. The proposed model contains three main extensions to the standard LSTM: First, an attention gated mechanism to capture long-term user preference changes. Second, a higher order interaction layer to alleviate data sparsity. Third, time aware LSTM cell gates to capture irregular time intervals between user interactions. We illustrate our solution using auxiliary information from Twitter and Google Plus to improve recommendations on YouTube. Extensive experiments show that the proposed model consistently outperforms state-of-the-art in terms of accuracy, diversity and novelty.

## 1 Introduction

Recommender systems are essential tools to successfully manage the information overload problem on Online Social Networks (OSNs) such as Facebook, Twitter and YouTube. The goal of a recommender system is to learn user preferences from historical interactions (e.g., likes, clicks and ratings), and automatically recommend interesting items to users. However, the limited user interactions on a single OSN are often insufficient to comprehensively capture user preferences (data sparsity problem) [He and Chua, 2017]. With the emergence and popularity of OSNs, users simultaneously engage in disparate OSNs. For example, a user may interact with his friends on Facebook, watch favorite TV shows on YouTube and share photos on Instagram. User activities performed on these OSNs represent his preferences from diverse perspectives. Cross-network recommender systems aggregate such interactions from multiple networks to create holistic user profiles and comprehensively capture user preferences [Yan *et al.*, 2014]. Despite the growing success of cross-network recommender systems, we identified two main limitations that degrade overall recommender quality.

(1) Existing cross-network solutions are based on linear models – The single network-based recommender literature has inspired many of the current cross-network solutions to use linear models such as Matrix Factorization (MF) [Yan *et al.*, 2014; 2015; Wanave and Takale, 2016]. However, linear models are unable to capture underlying complex relationships in user interactions. Therefore, we argue that the performance achieved by existing linear cross-network models are limited.

(2) Existing cross-network solutions cannot be used to capture the dynamic nature of online data – User preferences are subject to constant change over time. For example, during the FIFA world cup, a user may develop a new interest towards football related videos on YouTube. However, after the football season, he may no longer have such high interest, and would continue to watch other videos. Therefore, to provide timely recommendations, the continuous stream of new interactions need to be captured, and the underlying model parameters should be effectively updated to reflect his new interests (*online learning*). Despite successful attempts to develop single network-based online solutions [Devooght *et al.*, 2015; He *et al.*, 2016], such solutions are scarce for cross-network environments. Furthermore, we believe online solutions are more crucial for cross-network environments since user preference dynamics from multiple source network streams need to be incorporated to provide timely recommendations in the target network.

To overcome the above limitations, we propose a novel Long Short-Term Memory (LSTM) based online cross-network solution. The proposed method is able to model non-linear relationships in user interactions in an online setting. However, we argue that the use of standard LSTM limits the achievable recommender performance due to the following reasons.

First, the inability to handle highly sparse data – the standard LSTM is not designed to handle highly sparse data found in recommender environments.

Second, the practical inability to maintain long-range dependencies in user data – LSTMs encode the full user history into a fixed length cell vector to infer current preferences. Although encoding long-range dependencies in user data to a fixed-length cell vector is theoretically feasible, it was found to be challenging in practice (*cell state bottleneck*).

Third, the inability to handle irregular time gaps between interactions – LSTMs implicitly assume uniform time gaps between elements in the input sequence. However, user interactions occur at irregular time intervals, and varying time gaps convey important information. For example, when the time gaps between interactions are small, there is less chance for drastic user preference changes. Therefore, previous preferences become more relevant for current user preferences, and vice versa.

To overcome the above three limitations, we introduced a novel multi-layered LSTM model with – a higher-order interaction layer to handle data sparsity, a novel attention mechanism to reduce the burden of encoding the entire user history into a cell vector, and time aware input and forget gates to handle irregular time gaps between input interactions. The proposed model is then used for online cross-network recommendations. We demonstrate our solution using Twitter and Google Plus as source networks and YouTube as the target network. The proposed layered model also uses a cross-network topical layer to handle the heterogeneity of various cross-network data. Thus, the solution is general and can be easily extended to incorporate more source networks as needed.

In this paper, we first provide the preliminaries for LSTMs and Factorization Machines (FMs), and formulate the online cross-network recommender problem. Second, we present the proposed LSTM solution detailing the layered structure. Third, we compare our solution against multiple baselines and measure its effectiveness in terms of accuracy, diversity and novelty. We summarize our main contributions as follows:

- To the best of our knowledge, this is the first attempt to develop an online cross-network recommender solution.
- We propose a novel multi-layered LSTM model with a higher-order interaction layer, attention mechanism, and time aware input and forget gates to extend the standard LSTM to better support recommendations.
- We conduct extensive experiments to demonstrate that our method consistently outperforms state-of-the-art single and cross-network methods.

## 2 Related Work

**Cross-network recommender systems:** Unlike single network based solutions, cross-network solutions integrate user interactions from multiple source networks and create rich user profiles to conduct recommendations in a target network. Cross-network solutions consistently outperform single network solutions since they are more robust against data sparsity and cold start problems [Yan *et al.*, 2015; Wanave and Takale, 2016]. Recent practices in big data also suggest that more data beat better algorithms.

**Neural Networks (NNs) for recommender systems:** Compared to linear models, NN models are able to learn underlying complex relationships in user interactions. Most of the early recommender solutions used NNs only for feature processing [He and Chua, 2017]. Later models, better utilized the power of NNs by developing solutions with integral neural structures [Wang *et al.*, 2015; Salakhutdinov *et al.*, 2007]. Multiple solutions were developed based on various deep learning concepts such as multi-layer feed-forward networks [Xue *et al.*, 2017] and auto-encoders [Li *et al.*, 2015]. In contrast to these single-network based solutions, a recent deep NN based cross-network solution was proposed by [Elkahky *et al.*, 2015]. However, the solution is offline, and is unable to capture user preferences dynamics.

**Online recommender systems:** To provide timely recommendations, practical recommender solutions should model user preference dynamics with incoming data. Various single-network based online solutions such as MF based [He *et al.*, 2016; Devooght *et al.*, 2015], HMM-based [Kim *et al.*, 2014], and graph-based approaches were proposed to incrementally update model parameters using incoming data. However, these models were not designed to capture underlying non-linear relationships in user interactions. Recent online solutions based on Recurrent Neural Networks (RNNs) [Wu *et al.*, 2016; Tan *et al.*, 2016], and LSTM based models [Devooght and Bersini, 2017; Zhu *et al.*, 2017] easily outperformed other linear offline models. However, none of these models were cross-network solutions.

## 3 Model Preliminaries

### 3.1 Long Short-Term Memory (LSTM) Networks

The proposed model should be able to (1) capture non-linearities in user interactions, and (2) model user preference dynamics from online data. Therefore, we selected LSTMs as the basis for our model. They are a class of neural networks which are able to capture complex non-linear relationships in data, and model long-term dependencies in sequential data.

LSTMs were proposed to mitigate the popular vanishing or exploding gradient problem in RNNs. They were successfully used in many application areas, e.g., machine translation, speech recognition and handwriting recognition. LSTMs contain a chain of repeating modules with a cell state in each module to store important information at each input instance. The information flow through the memory cell is moderated by three gates: input gate (controls the input from the previous state), forget gate (controls the information to be ignored from the previous state), and output gate (computes the output from the current memory cell to perform predictions at the corresponding input instance).

Although LSTMs have recently been applied to the recommender environment [Devooght and Bersini, 2017], we argue that the standard LSTM is not directly applicable due to the following reasons: (1) inability to handle highly sparse data, (2) practical inability to maintain long-range dependencies in user data, and (3) inability to handle irregular time gaps between interactions. Our experiments further support this argument since the proposed LSTM based model consistently outperformed the standard LSTM.

### 3.2 Factorization Machines (FMs)

Sparse user-item interaction matrices in recommender systems create high dimensional and sparse feature vectors (user interaction vectors). To develop effective Machine Learning (ML) models for such highly sparse data, interactions between features need to be considered [Blondel *et al.*, 2016]. Successful solutions have been developed based on manually crafted combinatorial features. However, such solutions cannot be generalized for feature combinations that are not found in the training data, and conducting predictions for such unseen combinations lead to lower performance.

A recently proposed ML method, Factorization Machines (FMs) [Rendle, 2010] was designed to automatically learn feature interactions from raw data. FMs compute latent embeddings for each input feature and model feature interactions using element-wise inner products of the embedding vectors. Representing high-dimensional sparse features in a low-dimensional latent space allows the model to be generalized for unseen feature combinations [He and Chua, 2017]. We utilized FMs to mitigate data sparsity in the cross-network recommender setting since FMs have been proven to provide effective predictions for highly sparse data.

Given a real valued feature vector $\boldsymbol{x} \in \mathbb{R}^n$, FM estimates the target $\hat{y}(\boldsymbol{x})$ by modeling all interactions between each pair of features using factorized interaction parameters as follows:

$$\hat{y}(\boldsymbol{x}) = w_0 + \sum_{i=1}^{n} w_i x_i + \sum_{i=1}^{n} \sum_{j=i+1}^{n} \langle \boldsymbol{v_i}, \boldsymbol{v_j} \rangle \cdot x_i x_j \quad (1)$$

where $w_0$ is a global bias, $w_i$ is the interaction between the target and the feature $i$. The term $\langle \boldsymbol{v_i}, \boldsymbol{v_j} \rangle$ is the dot product between embedding vectors $\boldsymbol{v_i}, \boldsymbol{v_j} \in \mathbb{R}^k$, where $k$ is the dimensionality of the latent embedding space.

### 3.3 Problem Formulation

We represent each YouTube user $u \in U = [v_u^1, \ldots, v_u^{t-1}]$, as a sequence of his YouTube interactions, where $v_u^t$ is the interacted video at time $t$, and $T = [1, \ldots, t-1]$ are the timestamps of his interactions. At time $t$, the goal is to determine his current preferences on the target network by updating his previous preferences at $t-1$ using his *latest interactions* on source networks (i.e., interactions between $t$ and $t-1$). We formulated video recommendation as a Top-K recommender task, where at each timestamp, we predict a set of $K$ items that the user would most likely interact with. In the online setting, this process is continuously carried out with incoming interactions.

## 4 Methodology

The proposed method performs online cross-network recommendations using four layers (see Figure 1). The first two layers – cross-network topical layer and embedding layer form the inputs to our model. The third layer performs higher-order interactions to alleviate input sparsity. The fourth layer contains a modified LSTM cell with the novel attention mechanism and time aware input and forget gates to perform online recommendations (see Figure 2).

### 4.1 Cross-network Topical Layer

One of the main challenges in cross-network data is the heterogeneity of user interactions (e.g., user tweets, Google Plus posts and liked YouTube videos). The lack of explicit correlations between user interactions from different networks prevent their direct comparison and integration. Hence, we used a topic modeling approach to transfer and integrate interaction data from heterogeneous source networks to a homogeneous cross-network topical space. We assumed each interaction (i.e., tweet or Google Plus post) is associated with several topics and extracted them by analyzing textual data of the interactions. We considered each incoming tweet and post as a document and used Twitter-Latent Dirichlet Allocation (Twitter-LDA) [Zhao *et al.*, 2011] for topic extraction since it is more effective against short and noisy contents. Thus, each user interaction was represented as a topical distribution in a low dimensional cross-network topical space.

In an online setting, user preferences on the target network should be continuously updated. We used *latest interactions* on source networks as they reflect most recent user preferences [Yan *et al.*, 2015]. For a given user, consider $T = \{1, \ldots, t-1, t\}$ as timestamps of his target network interactions. At time step $t$, his *latest interactions* on source networks are between $t-1$ and $t$ ($\Delta t$). Hence, at each time step, we used corresponding *latest interactions* from both source networks to update his preferences on the target network.

The *latest interactions* on source network $S^a$ are encoded as a topical distribution vector $\boldsymbol{x^{a,t}} = \{x_1^{a,t}, \ldots, x_{K^t}^{a,t}\} \in \mathbb{R}^{K^t}$ where $K^t$ is the number of topics and $x_c^{a,t}$ is the frequency of topic $c$ (i.e., preference level towards the topic). Similarly, $\boldsymbol{x^{b,t}}$ encodes the *latest interactions* on source network $S^b$. The distributions from all source networks, $\boldsymbol{x^t} = \{\boldsymbol{x^{a,t}}; \boldsymbol{x^{b,t}}\} \in \mathbb{R}^{2K^t}$ form the output from the cross-network topical layer.

In addition to handling data heterogeneity, topic modeling also provides following advantages: Extendability – allows other source networks to be easily integrated, dimensionality reduction – reduces the overall complexity of the higher-order interaction layer (see Section 4.3), and interpretability – allows current and previous contexts to be easily compared, which is the basis of the proposed attention mechanism (see Section 4.4).

### 4.2 Embedding Layer

The outputs from the cross-network topical layer represent *latest interactions* of the user. We also used a one-hot encoded vector to uniquely identify the user. Together, they form the input features to our model. The computed topical distribution vector ($\boldsymbol{x^t}$) becomes sparse when $K^t$ is set to a sufficiently large value to capture finer level user interests, and/or when $\Delta t$ is small. Also, the one-hot encoded user vector contains a single non-zero value, and the sparsity increases with the number of users. Thus, the resulting sparse input features degrade prediction accuracy (data sparsity problem), since the training data does not contain a sufficient number of interactions for each input feature [Blondel *et al.*, 2016].

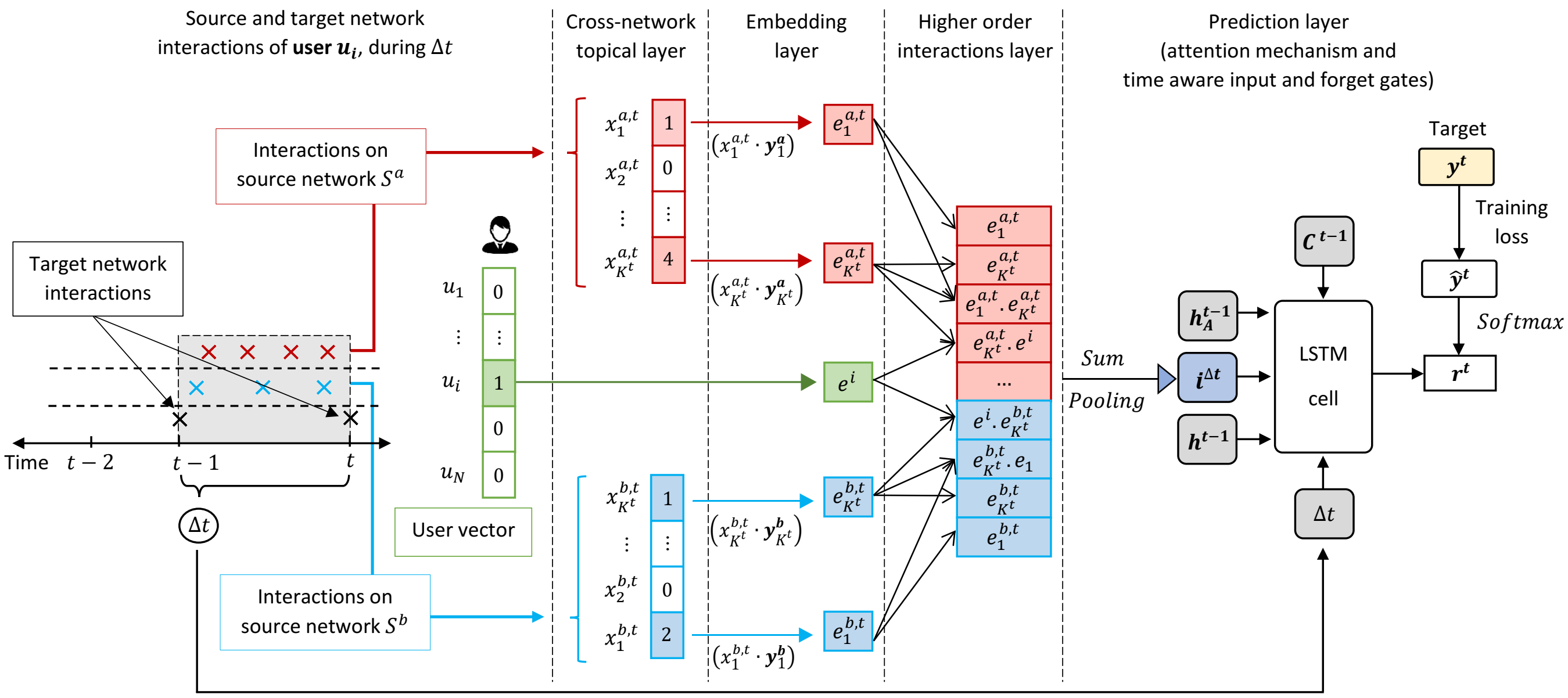

Figure 1: Proposed Model Architecture.

Inspired by FMs, the embedding layer projects each element (topic) in $\boldsymbol{x^t}$ to a dense vector representation with $k$ factors. Given the topical distribution $\boldsymbol{x^{a,t}} = \{x_1^{a,t}, \ldots, x_{K^t}^{a,t}\} \in \mathbb{R}^{K^t}$ for $S^a$, the model learns corresponding embeddings $E^{a,t} = \{\boldsymbol{e_1^{a,t}}, \ldots, \boldsymbol{e_{K^t}^{a,t}}\} \in \mathbb{R}^{k \times K^t}$ where $\boldsymbol{e_c^{a,t}} = x_c^{a,t} \cdot \boldsymbol{y_c^a} \in \mathbb{R}^k$, and $\boldsymbol{y_c^a} \in \mathbb{R}^k$ is the latent factorization of topic $c$. Note that we only need to learn embedding vectors for non-zero elements since $x_c^{a,t} = 0$ leads to $\boldsymbol{e_c^{a,t}} = 0$. Therefore, for $n$ non-zero elements, the resulting embedding vectors are $E^{a,t} \in \mathbb{R}^{k \times n}$. Similarly, $E^{b,t} \in \mathbb{R}^{k \times m}$ are embeddings learnt for $m$ non-zero elements in $\boldsymbol{x^{b,t}}$. We also learn a separate embedding $e^i \in \mathbb{R}^k$ for each user from the one-hot encoded user vector. The final output from the embedding layer is a concatenated embedding matrix $E_1^t = \left(E^{a,t} | E^{b,t} | e^i\right) \in \mathbb{R}^{k \times (n+m+1)}$, which is used by the higher-order interaction layer to alleviate sparsity (see Section 4.3).

### 4.3 Higher-Order Interaction Layer

We calculated second order interactions from the embeddings in $E_1^t$ as follows. For any pair of embeddings $\boldsymbol{e_i}, \boldsymbol{e_j} \in E_1^t$, a new vector $e_{ij} \in \mathbb{R}^k$ is computed using the element-wise product of the embeddings, where $(e_i \odot e_j)_k = e_{ik} \times e_{jk}$. Hence, $(n+m+1)$ embedding vectors are expanded to a matrix $E_2^t$ with $(n+m+1)(n+m)/2$ vectors. The output from the layer contains both first and second order interactions as $E^t = \left(E_1^t | E_2^t\right)$. Note that the layer can support other higher-order interactions ($> 2$). However, similar to the original work in FMs, we limit the degree of interactions to two, since second order interactions recorded a considerable performance gain under sparsity. Further performance improvements from other higher-order interactions can be explored as future work.

We used the popular sum pooling technique to convert the set of higher-order embeddings (column vectors in $E^t$) to a single vector as follows:

$$\boldsymbol{i^{\Delta t}} = \sum_{j=1}^{(n+m+1)(n+m)/2} \boldsymbol{E_j^t} \tag{2}$$

where $\boldsymbol{E_j^t}$ is column $j$ in $E^t$. The resultant $\boldsymbol{i^{\Delta t}} \in \mathbb{R}^k$ vector represents the specific user and his *latest interactions* on source networks, which forms the input to the LSTM at current time step $t$. Note that the sum pooling technique acts as a down sampling technique to reduce the number of parameters in the upper neural layers.

Standard FMs are a class of linear and shallow models since higher-order interactions are directly used for predictions. In contrast, our model mitigates such limitations by exposing the higher-order interactions to neural layers in the LSTM cell.

### 4.4 Attention Mechanism

At each time step, the attention mechanism allows the model to refer back to different sections of the user history. Therefore, it eases the burden on the LSTM cell to encode the full user history in the cell state vector. However, all parts of the user history are not equally relevant for recommendations in the current time step (e.g., due to seasonal preferences). Hence, the model computes *attention scores* to weigh the relevances of different parts of the user history.

**Attention scores calculation:** *Latest interactions* on source networks ($\boldsymbol{x^t}$) closely reflect current user preferences on the target network. Therefore, $\boldsymbol{x^t}$ also represents the current user context. The relevance of historical preferences are determined by comparing the current context ($\boldsymbol{x^t}$) to previous contexts $\{\boldsymbol{x^1}, \ldots, \boldsymbol{x^{t-1}}\}$. At current time step $t$, the attention score $\alpha_t^{t'}$ for learnt historical preferences at each pre-

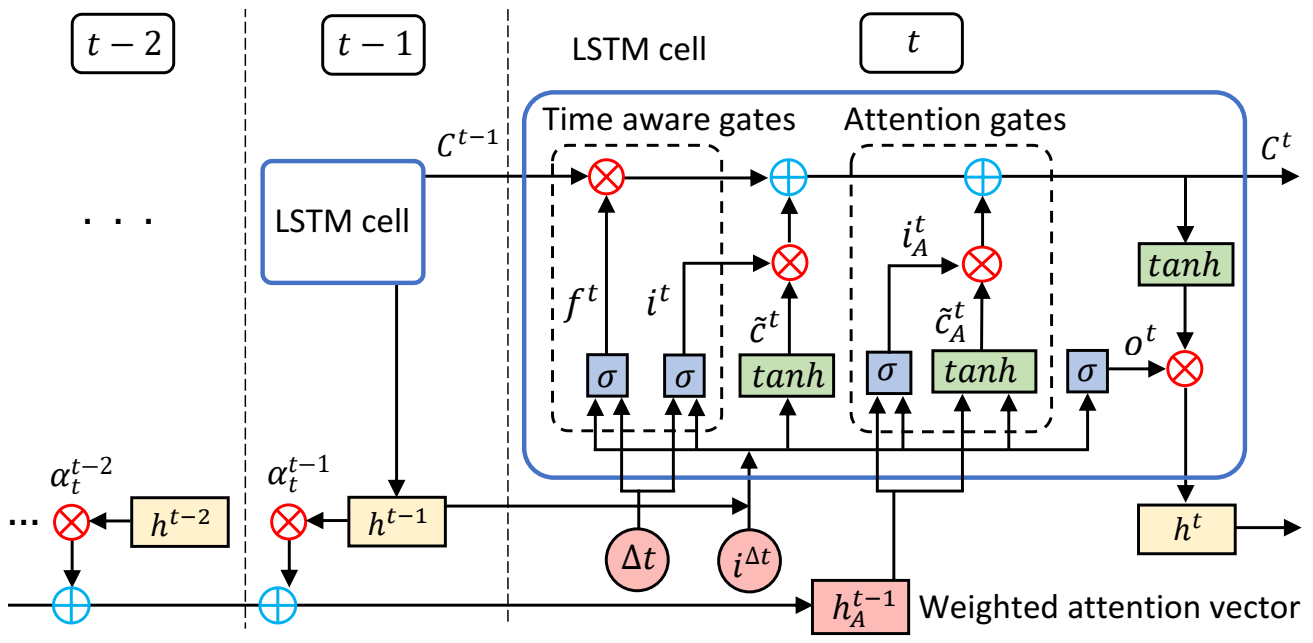

Figure 2: Modified LSTM Cell with Attention Mechanism and Time Aware Input and Forget Gates.

vious timestamp $t' \in \{1, \dots, t-1\}$ is calculated as follows:

$$\alpha_t^{t'} = \frac{\exp\left(sim(\boldsymbol{x^t}, \boldsymbol{x^{t'}})\right)}{\sum_{t'=1}^{t-1} \exp\left(sim(\boldsymbol{x^t}, \boldsymbol{x^{t'}})\right)},$$
$$sim(\boldsymbol{x^t}, \boldsymbol{x^{t'}}) = \frac{\sum_{c=1}^{2K^t} \boldsymbol{x^t_c} \cdot \boldsymbol{x^{t'}_c}}{\sqrt{\sum_{c=1}^{2K^t} \left(\boldsymbol{x^t_c}\right)^{\boldsymbol{2}}} \cdot \sqrt{\sum_{c=1}^{2K^t} \left(\boldsymbol{x^{t'}_c}\right)^{\boldsymbol{2}}}} \tag{3}$$

where $sim(\boldsymbol{x^t}, \boldsymbol{x^{t'}})$ is a cosine similarity function that computes a scalar importance value for $\boldsymbol{h^{t'}}$. Note that the function $sim(\boldsymbol{x^t}, \boldsymbol{x^{t'}})$ can also be learnt, for example, using a feed forward neural network to capture underlying complex similarities. We leave this extension as future work.

For a given user, consider an input sequence $I^t = \{\boldsymbol{i^{\Delta 1}}, \dots, \boldsymbol{i^{\Delta t}}\} \in \mathbb{R}^{k \times t}$ over $t$ timestamps, and corresponding LSTM output vectors over previous $t-1$ time steps as $H^{t-1} = \{\boldsymbol{h^1}, \dots, \boldsymbol{h^{t-1}}\} \in \mathbb{R}^{h \times (t-1)}$ where $h$ is the size of the output vector. At time step $t$, the goal is to compute the relevance of the above vectors to determine current preferences.

The information from previous outputs $(H^{t-1})$ are then aggregated based on their relevances to the current time step as a weighted attention vector $\boldsymbol{h_A^{t-1}} \in \mathbb{R}^h$ as follows:

$$\boldsymbol{h_A^{t-1}} = \sum_{t'=1}^{t-1} \alpha_t^{t'} \cdot \boldsymbol{h^{t'}} \tag{4}$$

The additional input $\boldsymbol{h_A^{t-1}}$ eases the burden on the LSTM cell to maintain the entire user preference history. We used two additional gates to effectively update the current cell vector using the relevant historical preferences in $\boldsymbol{h_A^{t-1}}$ as follows.

First, analogous to the input gate in the standard LSTM, an *input attention gate layer* $\boldsymbol{i_A^t} \in \mathbb{R}^h$ analyzes the current input $(\boldsymbol{i^{\Delta t}})$ and $\boldsymbol{h_A^{t-1}}$ to determine which values to be updated in the current cell state $(\boldsymbol{C^t})$. Specifically, $\boldsymbol{i_A^t}$ uses a sigmoid activation function $\sigma(x) = (1 + e^{-x})^{-1}$ to compute values between [0, 1], where each value determines the portion of an element in the cell state to be updated.

Second, analogous to the input modulation gate in the standard LSTM, a *modulation attention gate layer* $\boldsymbol{\tilde{c}_A^t} \in \mathbb{R}^h$ also analyzes the current input $(\boldsymbol{i^{\Delta t}})$ and $\boldsymbol{h_A^{t-1}}$. Then, computes a set of candidate values based on previous preferences that could be added to update the current cell state $(\boldsymbol{C^t})$. Specifically, $\boldsymbol{\tilde{c}_A^t}$ uses a hyperbolic tangent activation function $\tanh(x) = (e^x - e^{-x})(e^x + e^{-x})^{-1}$ to compute candidate values between [-1, 1] for each element in the cell state.

The final result $\boldsymbol{c_A^t} \in \mathbb{R}^h$ from the attention mechanism is obtained by scaling the initial candidate values $(\boldsymbol{\tilde{c}_A^t})$ by the portions of elements to be updated $(\boldsymbol{i_A^t})$. This vector represents the most relevant historical preferences for the user at the current time step.

$$\begin{aligned} \boldsymbol{i_A^t} &= \sigma(W_A^i \boldsymbol{i^{\Delta t}} + U_A^i \boldsymbol{h_A^{t-1}} + \boldsymbol{b_A^i}) \\ \boldsymbol{\tilde{c}_A^t} &= \tanh(W_A^c \boldsymbol{i^{\Delta t}} + U_A^c \boldsymbol{h_A^{t-1}} + \boldsymbol{b_A^c}) \\ \boldsymbol{c_A^t} &= \boldsymbol{i_A^t} \odot \boldsymbol{\tilde{c}_A^t} \end{aligned} \tag{5}$$

where $W_A^i, W_A^c \in \mathbb{R}^{h \times k}$ and $U_A^i, U_A^c \in \mathbb{R}^{h \times h}$ are latent linear transformation parameters, and $\boldsymbol{b_A^i}, \boldsymbol{b_A^c} \in \mathbb{R}^h$ are bias terms. Hence, the cell state is updated by adding these values (see equation 7).

In an online setting, the model is trained offline and conducts continuous predictions using incremental updates. However, due to dynamics in the environment, recommender accuracy tends to drop over time. Our experiments show that by effectively incorporating previously learnt outputs, the attention mechanism helps maintain the achieved recommendation quality.

## 4.5 Time Aware Input and Forget Gates

The LSTM cell states are updated at each time step to capture timely user preferences. However, when the time gap from the previous interaction increases, the changes in user preferences also tend to increase. Accordingly, the contribution from the previous cell state should be less. Therefore, based on the time gap between interactions, the proposed forget gate models a temporal decay to discard more information from the previous state. Simultaneously, the contribution from the current input $(\boldsymbol{i^{\Delta t}})$ should increase since, *latest interactions* closely reflect current preferences. We modeled the proposed input gate $\boldsymbol{i^t} \in \mathbb{R}^h$ and forget gate $\boldsymbol{f^t} \in \mathbb{R}^h$ as follows:

$$\begin{aligned} \boldsymbol{i^t} &= (1 - e^{(-\Delta t)}) \cdot \sigma(W^i \boldsymbol{i^{\Delta t}} + U^i \boldsymbol{h^{t-1}} + \boldsymbol{b^i}) \\ \boldsymbol{f^t} &= e^{(-\Delta t)} \cdot \sigma(W^f \boldsymbol{i^{\Delta t}} + U^f \boldsymbol{h^{t-1}} + \boldsymbol{b^f}) \end{aligned} \tag{6}$$

where $W^i, W^f \in \mathbb{R}^{h \times k}$ and $U^i, U^f \in \mathbb{R}^{h \times h}$ are latent linear transformation parameters, and $\boldsymbol{b^i}, \boldsymbol{b^f} \in \mathbb{R}^h$ are bias terms.

In addition to these gate operations, at each time step $t$, the LSTM cell uses the following gate operations to calculate the current cell state output $(\boldsymbol{h^t})$ as follows:

$$\begin{aligned} \boldsymbol{o^t} &= \sigma(W^o \boldsymbol{i^{\Delta t}} + U^o \boldsymbol{h^{t-1}} + \boldsymbol{b^o}) \\ \boldsymbol{\tilde{c}_t} &= \tanh(W^c \boldsymbol{i^{\Delta t}} + U^c \boldsymbol{h^{t-1}} + \boldsymbol{b^c}) \\ \boldsymbol{C^t} &= \boldsymbol{f^t} \odot \boldsymbol{C^{t-1}} + \boldsymbol{i^t} \odot \boldsymbol{\tilde{c}^t} + \boldsymbol{i_A^t} \odot \boldsymbol{\tilde{c}_A^t} \\ \boldsymbol{h^t} &= \boldsymbol{o^t} \odot \tanh(\boldsymbol{C^t}) \end{aligned} \tag{7}$$

where $\boldsymbol{o^t} \in \mathbb{R}^h$ is the output gate, $\boldsymbol{\tilde{c}_t} \in \mathbb{R}^h$ is the input modulation gate, $W^o, W^c \in \mathbb{R}^{h \times k}$ and $U^o, U^c \in \mathbb{R}^{h \times h}$ are latent linear transformation parameters, and $\boldsymbol{b_o}, \boldsymbol{b_c} \in \mathbb{R}^h$ are bias terms. Note that these operations are analogous to the standard LSTM gate operations. Additionally, the proposed time aware input and forget gates, and the supplementary input from the attention mechanism are used to calculate the cell state vector $\boldsymbol{C^t}$.

### 4.6 Prediction Layer

The LSTM cell output ($\boldsymbol{h^t}$) represents current user preferences. We used a neural layer with a hyperbolic tangent activation function ($tanh$) to transform $\boldsymbol{h^t}$ in to a prediction vector $\boldsymbol{r^t} \in \mathbb{R}^I$, where each element $r_i^t \in \boldsymbol{r^t}$ is a scalar value which weighs the possibility of interacting with item $i \in I$. Finally, a softmax function normalizes these weights, and outputs the probability of interacting with each item in vector $\boldsymbol{\hat{y}^t} \in \mathbb{R}^I$.

$$\boldsymbol{r^t} = tanh(W^r \boldsymbol{h^t} + \boldsymbol{b}^r)$$
$$\boldsymbol{\hat{y}^t} = softmax(\boldsymbol{r^t}) = \frac{\exp(r_i^t)}{\sum_{i=1}^{I} \exp(r_i^t)} \tag{8}$$

where $\boldsymbol{\hat{y}^t}$ is the final prediction, $W^r \in \mathbb{R}^{I \times h}$ is a latent linear transformation parameter, and $\boldsymbol{b}^r \in \mathbb{R}^I$ is a bias term.

In line with online recommendation literature, we limited the recommended list to the 100 highest ranked items (Top-100) [He *et al.*, 2016]. Hence, from the prediction vector ($\boldsymbol{\hat{y}^t}$) we selected the 100 items with the highest values.

**Learning:** The target (actual) interaction is represented as a one-hot encoded vector $\boldsymbol{y^t} \in \mathbb{R}^I$, which was used as the ground truth to evaluate model performance. Since each output element $\hat{y}_i^t \in \boldsymbol{\hat{y}^t}$ determines the probability of interaction with an item $i$, the loss function to minimize is computed by the cross-entropy error between $\boldsymbol{y^t}$ and $\boldsymbol{\hat{y}^t}$ distributions as follows:

$$\boldsymbol{L} = -\sum_{i=1}^{I} \boldsymbol{y_i^t} \log(\boldsymbol{\hat{y}_i^t}) + (1 - \boldsymbol{y_i^t}) \log(1 - \boldsymbol{\hat{y}_i^t}) \tag{9}$$

All model equations and the loss function are analytically differentiable. Therefore, the model can be readily trained from end-to-end. We used Adam optimization (ADAM) over the most common Stochastic Gradient Descent (SGD) algorithm for optimizations since, ADAM adaptively updates the learning rate during training, leading to faster convergence compared to the vanilla SGD.

**Preventing overfitting:** Deep neural networks tend to easily overfit the training data. Hence, we used the dropout technique, which is a widely used regularization technique to prevent overfitting in neural networks. During training, the dropout technique removes a randomly selected set of neurons and their connections. Therefore, on each parameter update, only a portion of model parameters that contribute to predicting $\boldsymbol{\hat{y}^t}$ is updated. This avoids forming complex co-adaptations of neurons on training data. Note that dropout was disabled for testing and the whole network was used for predictions.

**Online update:** In practice, the model should be initially trained offline on historical data and updated online from incoming interactions. Let $W, U$ and $\boldsymbol{b}$ denote the sets of model parameters learnt from offline training, and $\boldsymbol{y_n^t}$ an incoming target network interaction for a user. We perform optimization steps only for the new observation by making a prediction $\boldsymbol{\hat{y}_p^t}$ for the user, and backpropagating the error $(\boldsymbol{y_n^t} - \boldsymbol{\hat{y}_p^t})$ to update model parameters. However, from a global perspective, the new interaction should not excessively change the learned parameters. Hence, a stopping criterion is set to limit the number of optimization iterations for new interactions. However, our experiments show that very few iterations (1-3) are sufficient to obtain good results.

## 5 Experiments

**Dataset and setup:** We extracted users with Twitter, Google Plus and YouTube interactions from two public datasets [Lim *et al.*, 2015; Yan *et al.*, 2014], and scraped timestamped interactions over a 2-year period (1st March 2015 to 29th February 2017). We used Tweets and Google Plus posts as source network interactions, and YouTube videos either liked or added to playlists as target network interactions. The high sparsity in original datasets makes it difficult to evaluate recommendation algorithms. Therefore, in line with common practices, we filtered out users with less than 10 interactions on Twitter, 5 interactions on Google Plus, and users and videos with less than 10 interactions on YouTube. We used a lower threshold for Google Plus since comparatively it contained a lower number of interactions. The final dataset contained 2,536 users, 13,829 YouTube videos, and the overall sparsity of the user-video matrix was 99.64%. Across all networks, the average number of interactions for each user was 26. The time gaps between interactions ranged from zero seconds to 7 months, and the average time gap per user was 6.6 days.

We sorted both source and target network user interactions in chronological order. Then, from each user, the oldest 70% of target network interactions and source network interactions within the same time period were used as the training set. Similarly, the next 10% was used as the validation set (to tune hyper-parameters), and the latest 20% was used as the test set (held out for predictions). We simulated a dynamic data stream to create a realistic online recommender environment. For each user, we recommended a list of Top-K videos, where K was set to 100 similar to [He *et al.*, 2016]. Then, for each incoming user interaction, we calculated the Hit-Ratio (HR) by counting the number of times the Ground Truth items (GT) appeared in the corresponding list (hits) as follows:

$$HR@K = \frac{NumberofHits@K}{|GT|} \tag{10}$$

However, HR does not capture the raked position of a hit (recall-based metric). Therefore, we used the Normalized Discounted Cumulative Gain (NDCG), which assigns higher importance to top-ranked hits compared to lower ranked hits as follows:

$$NDCG@K = Z_K \sum_{i=1}^{K} \frac{2^{r_i} - 1}{\log_2(i+1)} \tag{11}$$

where a normalizer $Z_K$ is used to ensure that the perfect ranking has a value of 1, and $r_i$ is the weighted relevance of the item at position $i$. We used a simple binary relevance, where $r_i = 1$ if the item appears in the test set, and 0 if otherwise. We calculated each metric per user, and the final results were averaged across all users.

**Baselines:** We compared our model against popularity based, MF based, single network based, cross-network based models, and several variations of the proposed model.

- **TimePop:** Calculates the most popular $K$ items for a given time interval and recommends them to all users. We considered a day as the time interval and recommended the daily top-100 popular videos to all users.
- **DMF** [Devooght *et al.*, 2015]: Dynamic MF is the state-of-the-art single network-based MF method designed for online training. For model parameters, we used the suggested values in the original implementation[1].
- **TDCN** [Perera and Zimmermann, 2017]: Time-Dependent Cross-Network is an offline MF based linear model, which uses timestamped interactions across networks to conduct recommendations. For fair comparisons in an online setting, we retrained the model at biweekly intervals. The authors suggest that smaller time intervals increase performance. However, due to data sparsity, we used biweekly intervals, similar to the original work.
- **Time-LSTM** [Zhu *et al.*, 2017]: A single network-based LSTM model which incorporates timestamps of user interactions to capture short- and long-term user preferences. The authors proposed three variations of the model, and we used the best performing model (Time-LSTM3) as the baseline. Similar to our model, we used the simulated dynamic data stream to compare online performance.
- **CLSTM:** Due to the absence of online non-linear cross-network solutions in the literature, we formed a Cross-network LSTM solution, which is a variation of our model. Essentially, CLSTM directly uses the outputs from the cross-network topical layer as inputs (see Section 4.1) to the standard LSTM cell.

## 6 Discussion

**Prediction accuracy:** We plot the changes in accuracy (HR and NDCG) against the test interactions (see Figure 3). The number of test interactions varies among users. Therefore, the results were averaged among participating users. Note that the validation set was used to select the best performing version of the model, and the average across multiple runs were recorded. However, we observed a low-performance variance across runs ($\pm 1.3\%$).

TimePop has the lowest accuracy since it does not provide personalized recommendations. Compared to linear models (DMF and TDCN), non-linear models (Time-LSTM, CLSTM and Proposed) show higher accuracy since non-linear models are able to better capture underlying complex relationships in data. Although both linear models are based on MF, DMF shows comparatively low accuracy since it is based on a single network, while TDCN makes use of auxiliary information from multiple networks. Similarly, since Time-LSTM is based on a single network, it has the lowest performance compared to the other non-liner and cross-network models (CLSTM and Proposed). CLSTM outperforms all other baselines, which illustrates the benefits of non-linear cross-network models. However, the proposed model consistently outperforms CLSTM.

[1] https://github.com/rdevooght/MF-with-prior-and-updates

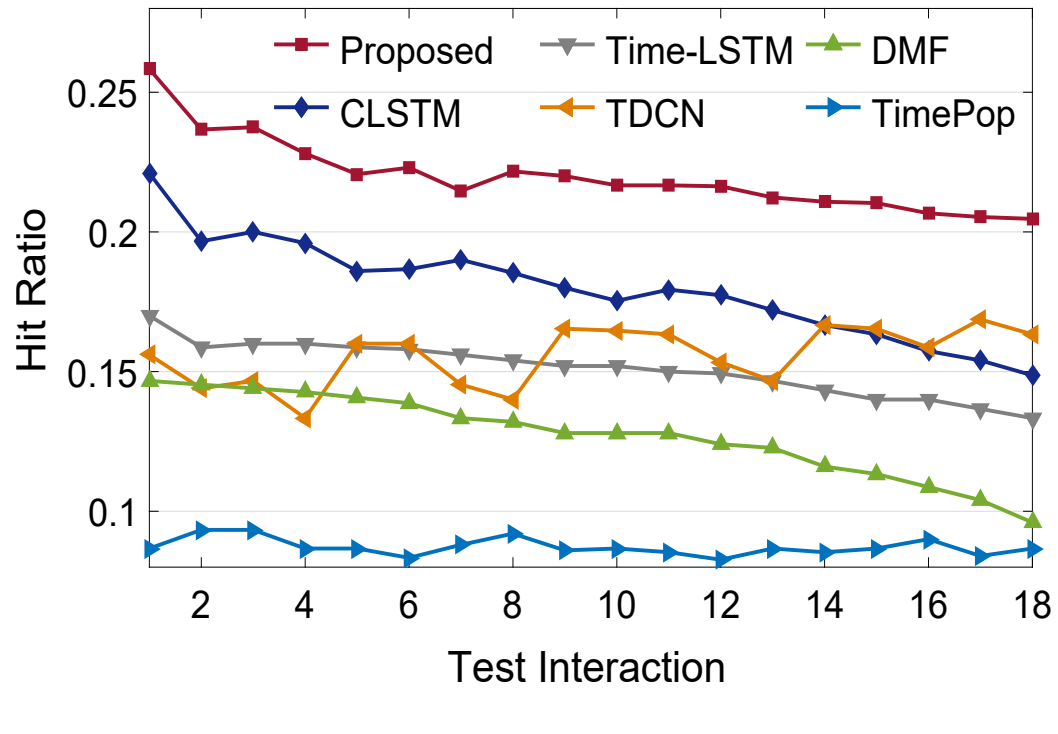

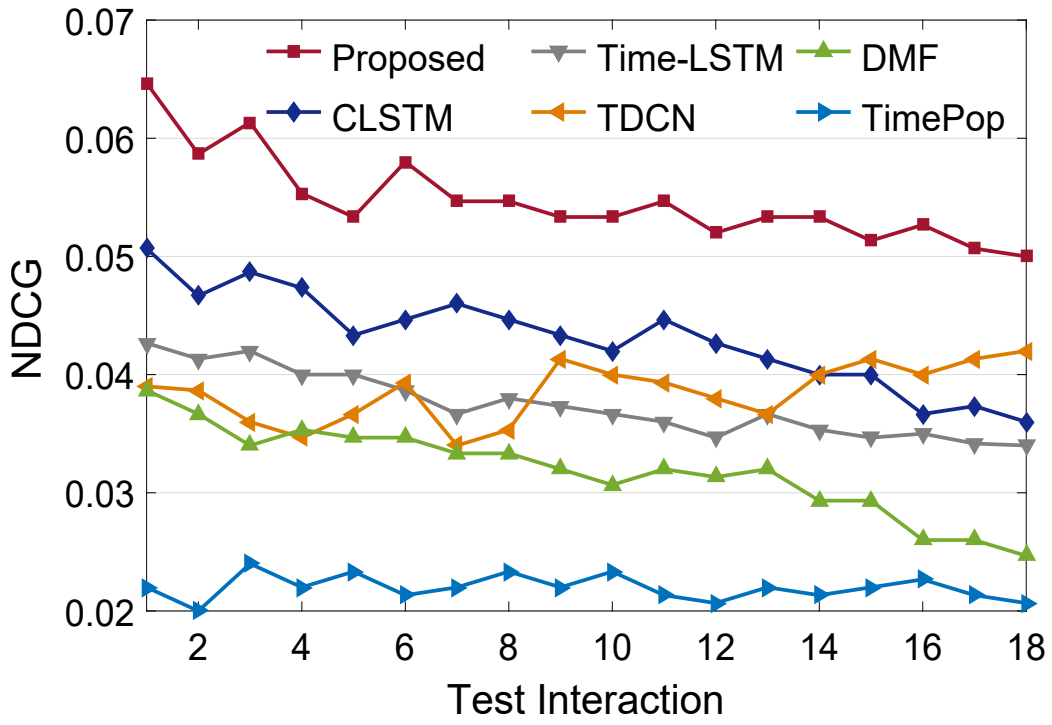

Figure 3: Online prediction accuracy over time.

Only TimePop and the proposed model are able to better maintain the achieved accuracy over time. TimePop is successful since it does not rely on a trained model, but is based on a simple statistic - the popularity of videos. The accuracy of TDCN increases over time since it is retrained at biweekly intervals using an increasingly large dataset. Note that since TDCN is retrained, it is not an online solution and therefore, cannot be used to compare the accuracy decline over time.

Furthermore, additional experiments showed that in all methods, the accuracy declines with higher top-K values due to increasing false positive rates. However, the proposed model outperformed all baselines.

**Impact from higher-order interaction layer:** We compared our modal against a modified version – NoHO, which does not contain the higher-order interaction layer. The proposed model consistently outperformed NoHO over time (see Figure 4). The higher-order interaction layer was introduced to mitigate data sparsity (or density). Therefore, to measure its effectiveness, we calculated density values (fraction of non-zero elements) for input features ($\boldsymbol{x}^t$) across all time intervals. We plotted average accuracy values against varying density values, for both the proposed and NoHO models (see Figure 5). When sparsity increases, both models de-creased in accuracy. However, the proposed model consistently achieved higher accuracy and is, therefore, more robust against sparsity.

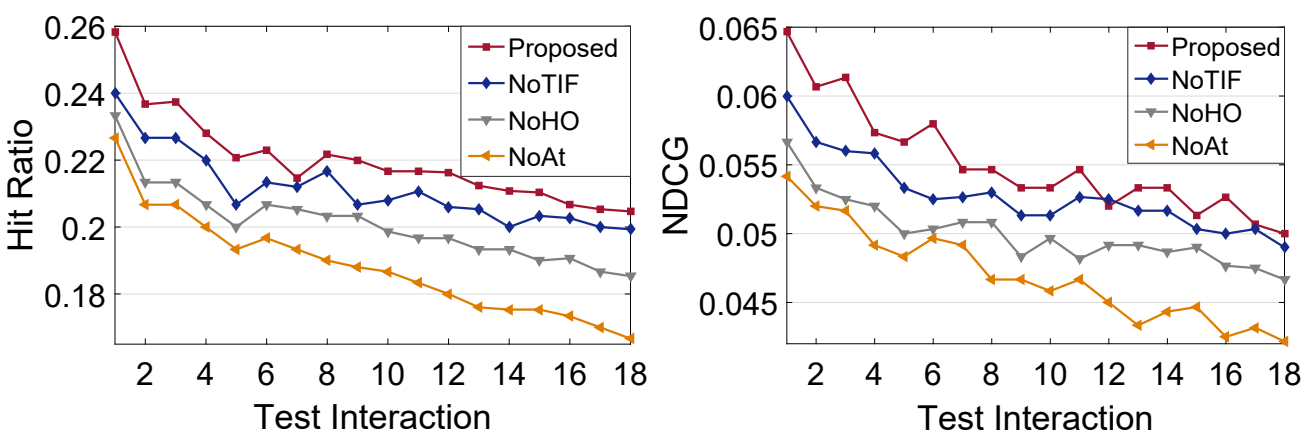

Figure 4: Effects of the proposed extensions on accuracy.

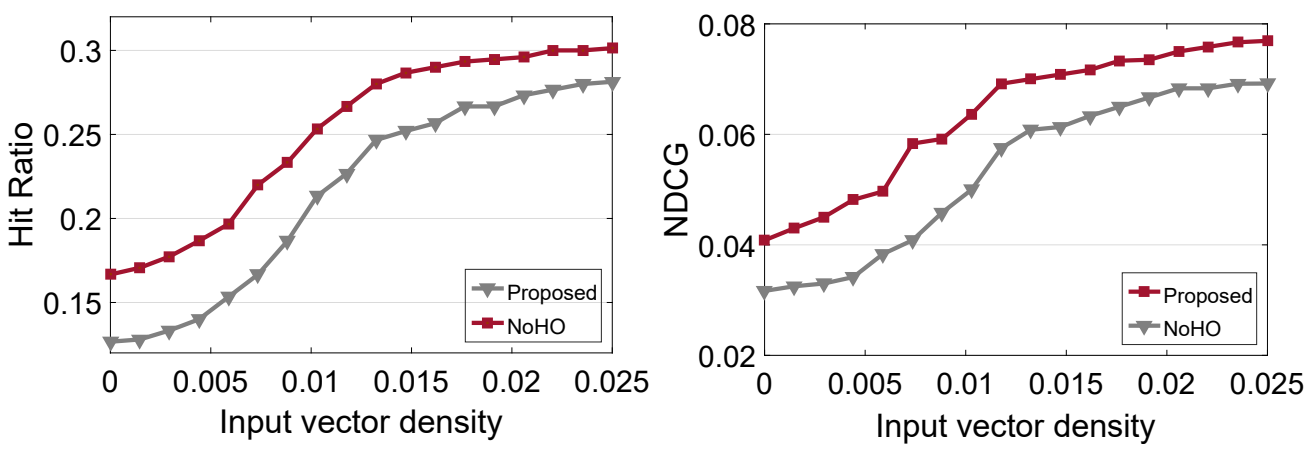

Figure 5: Effects of the higher-order interaction layer on accuracy.

**Impact from attention mechanism:** We compared the proposed modal against another modified version - NoAt, which does not contain the attention mechanism. The proposed model constantly achieved higher accuracies compared to NoAt (see Figure 4). When new interactions were streamed in, both models showed a drop in accuracy. However, the rate of accuracy decline is higher in NoAt, which shows the contribution from the attention mechanism to maintain the achieved accuracy over time.

**Impact from time aware input and forget gates:** We also compared the proposed modal against another modified version - NoTIF, which does not contain time aware input and forget gates. The proposed model maintained a higher accuracy compared to NoTIF (see Figure 4). However, the contribution to model accuracy from the time aware input and forget gates is comparatively low compared to other variations of the model. This could be improved by including more complex temporal functions.

**Impact from the dimensionality of embedding vectors:** We compared the changes in model accuracy against varying dimensionality ($k$) values in the embedding layer (see Figure 6). For each $k$ value, the accuracy results were averaged across time steps. When $k$ values increase, the representation power of input features increases, hence the model accuracy also increases. However, higher $k$ values could lead to overfitting. The optimum $k$ value may vary based on the dataset size since large datasets require higher $k$ values to comprehensively capture the finer level details.

**Diversity and Novelty:** Both HR and NDCG measured accuracy, which is insufficient to capture overall user satisfaction. For example, recommending similar videos (i.e., in terms of topic, author and genre), that covers a small fraction of user's preferences could still lead to higher accuracy, but the user could lose interest over time. Therefore, we calculated two additional metrics, diversity [Avazpour *et al.*, 2014] and novelty [Zhang, 2013] to evaluate the overall effectiveness of our solution. On average, compared to the closest CLSTM approach, the proposed model showed 7.7% and 12.8% improvements in diversity and novelty. Therefore, it is evident that the proposed model improves the overall recommender quality.

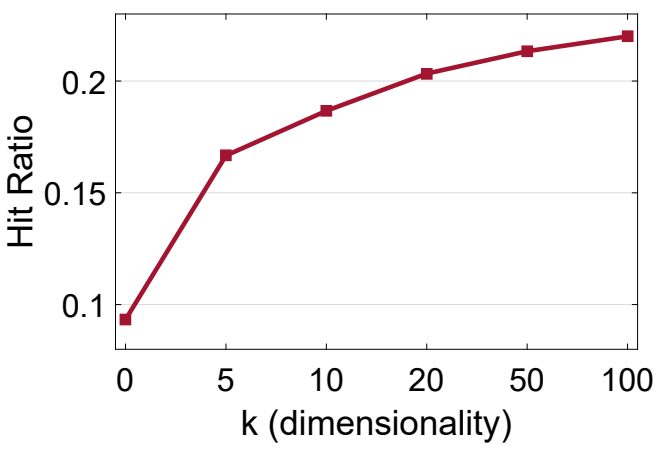

Figure 6: Prediction accuracy against the dimensionality of embedding layers.

**Model parameters**: We set the following parameter values, which were tuned using the validation set.

- We used a grid search algorithm to set the number of topics ($K^t$) to 60 and the final results were not highly sensitive to minor changes in $K^t$ ($\pm$10).
- We also used a grid search algorithm to set the number of dimensions in the embedding layer ($k$) to 100, the number of hidden units ($h$) to 400, and the dropout ratio to 0.35.
- The learning rate ($\mu$) was set to a fairly small value (0.001) to obtain the local minimum.

## 7 Conclusion and Further Work

We identified two main limitations in existing cross-network recommender solutions: (1) failure to capture underlying complex relationships in user interactions, and (2) failure to adapt to the dynamic recommender environment. Thus, to the best of our knowledge, we propose the first online cross-network recommender solution to mitigate these issues. The proposed multi-layered LSTM model introduced a higher-order interaction layer, attention mechanism, and time aware input and forget gates to extend the standard LSTM to better support recommendations. Our model consistently outperformed multiple baselines in terms of accuracy, diversity and novelty. Furthermore, as future work, we plan to extend our model to incorporate social influences on user preferences (e.g., changes in friends' preferences).

The proposed solution helps alleviate two significant limitations in cross-network recommendations, which we believe lays a foundation to make cross-network recommendation a reality in practice.

## Acknowledgments

This research was supported in part by the National Natural Science Foundation of China under Grant no. 61472266 and by the National University of Singapore (Suzhou) Research Institute, 377 Lin Quan Street, Suzhou Industrial Park, Jiang Su, People's Republic of China, 215123.